\documentclass{article}
\usepackage{mypreprint}
\usepackage{lipsum}

\begin{document}

\twocolumn[
\mytitle{SHAPE: An Unified Approach to Evaluate the Contribution and Cooperation of Individual Modalities}


\mysetsymbol{equal}{*}

\begin{myauthorlist}
\myauthor{Pengbo Hu}{equal,ustc1,bsbii}
\myauthor{Xingyu Li}{equal,bsbii}
\myauthor{Yi Zhou}{ustc2,bsbii}
% \myemail{\{pbhu, xyli1905, yzhou\}@bsbii.cn}
\end{myauthorlist}

\myaffiliation{bsbii}{Shanghai Center for Brain Science and Brain-Inspired Technology, Shanghai, China}
\myaffiliation{ustc1}{School of Mathematical Sciences, University of Science and Technology of China, Hefei, China}
\myaffiliation{ustc2}{National Engineering Laboratory for Brain-inspired Intelligence Technology and Application, School of Information Science and Technology, University of Science and Technology of China, Hefei, China}

\mycorrespondingauthor{Yi Zhou}{yizhoujoey@gmail.com}

\mykeywords{Multi-modal Learning, Shapley Value}

\vskip 0.3in
]


% The \myEqualContribution command is standard text for equal contribution.

\printAffiliationsAndNotice{\myEqualContribution} % otherwise use the standard text.

\begin{abstract}
As deep learning advances, there is an ever-growing demand for models capable of synthesizing information from multi-modal resources to address the complex tasks raised from real-life applications. Recently, many large multi-modal datasets have been collected, on which researchers actively explore different methods of fusing multi-modal information. However, little attention has been paid to quantifying the contribution of different modalities within the proposed models. 
In this paper, we propose the {\bf SH}apley v{\bf A}lue-based {\bf PE}rceptual (SHAPE) scores that measure the marginal contribution of individual modalities and the degree of cooperation across modalities. Using these scores, we systematically evaluate different fusion methods on different multi-modal datasets for different tasks. Our experiments suggest that for some tasks where different modalities are complementary, the multi-modal models still tend to use the dominant modality alone and ignore the cooperation across modalities. On the other hand, models learn to exploit cross-modal cooperation when different modalities are indispensable for the task. In this case, the scores indicate it is better to fuse different modalities at relatively early stages. We hope our scores can help improve the understanding of how the present multi-modal models operate on different modalities and encourage more sophisticated methods of integrating multiple modalities.
\end{abstract}

% main body

\section{Introduction}

The present deep learning algorithms thrive on a wide spectrum of real-life tasks, such as those in the fields of computer vision and natural language processing, for which the essence is to learn the underlying distribution from the dataset. 
Inspired by these achievements, researchers march on to conquer the more complex tasks that put emphasis on reasoning and knowledge.
One character of these complex tasks is that they normally involve data from distinct modalities. 
For example, in the Visual Question Answering (VQA) task, one must answer the questions based on the knowledge extracted from the corresponding image~\cite{goyal2017making,antol2015vqa}.
The ability to fuse multi-modal information serves as the cornerstone to developing solutions to those complex tasks,
and it presents severe challenges to the deep learning community.

Recently, many large multi-modal datasets have been collected, 
e.g.~\cite{vadicamo2017cross,zadeh2018multimodal,bowman2015large,liang_multibench_2021},
laying the ground for exploring sophisticated fusion strategies.
Beyond the naive methods of early and late fusion strategies, increasing attention has been paid to developing and practicing hybrid fusing strategies~\cite{wang2021survey,baltruvsaitis2018multimodal,wang_what_2020,bai_m2p2_2021}. Those efforts constantly push forward the model performance on the multi-modal datasets.
Despite the success, people gradually realize a subtle limitation of the present progresses: 
the model performance is almost solely judged on their prediction accuracy.
A higher accuracy does not necessarily translate to the better usage of multi-modal information.
For example, in the CMU-MOSEI datasets, the accuracy of a unimodal model trained on the text modality is very close to the ones of SOTA multi-modal models~\cite{zadeh2018multimodal,delbrouck2020transformer}.  
Consequently, there is an urgent need for a quantitative metric that can capture how the underlying models operate on the multi-modal information.

\footnotetext[1]{Implementation: https://github.com/zhouyilab/SHAPE.}

To evaluate the individual modalities' importance, \citet{gat2021perceptual} proposed the perceptual score, 
which estimates the contribution of a modality with its marginal performance gain at the presence of other modalities.
However, this score cannot measure cross-modal cooperation and, in the presence of more than two modalities,  
its attribution of the importance to individual modalities does not correctly reflect their marginal contribution.
To address the above issues, we draw inspirations from the game theory and propose the {\bf SH}apley v{\bf A}lue-based {\bf PE}rceptual (SHAPE) scores to directly evaluate the marginal contribution and the cooperation of individual modalities. 
The validity of the SHAPE scores roots in the properties of the celebrated Shapley value~\cite{shapley1953value}. 
With the SHAPE scores, we aim at providing additional perspectives for evaluating the performance of the multi-modal models, which in turn would help building sophisticated fusion strategies.

Our contributions are three-fold:
\begin{itemize}
    \item[1.] We propose the SHAPE scores, which
    are theoretically guaranteed to quantify the marginal contribution of individual modalities and the cross-modal cooperation. 
    \item[2.]  With the SHAPE scores, we systematically evaluate the usage of multi-modal information by models  
    using different fusion methods on the different multi-modal datasets for different tasks.
    \item[3.] We extend the perceptual score to a more informative fine-grained version. 
    Moreover, we demonstrate that the perceptual score implicitly measures the correlation among modalities within the dataset.
\end{itemize}

\section{Related Work}

\subsection{Multi-modal Fusion Method}
The fusion strategies can be roughly classified as early fusion, late fusion, and hybrid fusion~\cite{baltruvsaitis2018multimodal}. In the early fusion, one directly concatenates inputs from individual modalities to build a unified input embedding. In this case, a single pathway model is enough to make a prediction, making the training pipeline easier~\cite{kiela2019supervised}. The late fusion usually processes individual modalities in different pathways. 
The fusion operation is typically implemented on the output from each pathway. The later fusion strategy ignores the low-level feature interaction, making it hard to learn the joint representation of modalities. However, in practice, later fusion enables a model using the different pre-trained models for different pathways, making it more flexible than the early fusion~\cite{rodrigues2019multimodal}. Hybrid fusion has no specific recipe. The different modalities can be integrated at different stages, on different feature levels, and among different modules. It could effectively promote collaboration among individual modalities. Generally, the optimal hybrid fusion strategy depends on the type of the underlying task. For example, in the VQA task, the co-attention-based fusion methods are proved to be better than other strategies~\cite{liu2019densely}. 

\subsection{Modality Contribution Attribution}
Presently, most evaluation works focus on the feature importance. For example, \citet{lundberg2017unified}
proposed the SHapley Additive exPlanations (SHAP) value, which can estimate feature importance using an approximation of the features Shapley values. 
For evaluating modal importance, \citet{hessel2020does} adopted an indirect attribution method. They designed a tool called the empirical multimodally-additive function projection (EMAP), which is an effective modal that elimates the cross-modal cooperation in the reference multi-modal model. Through comparing the performance of the reference multi-modal model and its EMAP, one could estimate how much the cross-modal cooperation contribute.

The work most relevant to ours is the perceptual score, which is proposed by \citet{gat2021perceptual} to assess the importance of a modality in the input data for a given model. 
They adopted a permutation-based approach for removing the modality influence. Specifically, let $x^{M}$ denote the feature of target modality $M$ in the input data point. In order to remove $M$'s influence, $x^{M}$ is substituted with a randomly chosen one from another data point, i.e., $x'^{M}$. The resulting classifier on the permutated data is called the permuted classifier.
The perceptual score for $M$ is defined as the normalized difference between the accuracies of the normal classifier and the corresponding permutated classifier.

It is worth noting that the perceptual score does not account for the marginal contribution of the target modality $M$. Instead, it measures the $M$'s contribution conditioned on all other modalities in the dataset. The two types of contributions are the same only in the two-modal case. In the presence of more modalities, it requires a more sophisticated method to compute the marginal contribution. Moreover, there still lacks a quantitative method that isolates the cooperation between different modalities.

\subsection{Shapley Value}

The Shapley value~\cite{shapley1953value} was initially proposed in the coalitional game theory for fair attribution of payouts among the players depending on their contribution to the total payout.  
A multi-player game is usually defined as a real-valued utility function $V$ on its participants. 
Let $\mathcal{N}:=\{1, \dots, n\}$ denote the set of players in the game. 
The player can form different coalitions (i.e., subsets of $\mathcal{N}$) whose net effect can be either cooperative or competing. 
Consider a player $i$ and a subset $S\subseteq \mathcal{N}/\{i\}$, the marginal contribution of $i$ to $S$ can be evaluated by
\begin{equation*}
    m_i(S;V) = V(S\cup \{i\}) - V(S).
\end{equation*}
Since a player can interact with any other players, in order to evaluate the marginal contribution of $i$ in the game $V(\mathcal{N})$ one needs to average over all possible $m_i(S;V)$. This is exactly how the Shapley value $\phi_i$ is defined,
\begin{equation*}
    \phi_i(\mathcal{N};V) := \sum_{S\subseteq \mathcal{N}/\{i\}} \frac{|S|!(n-|S|-1)!}{n!}m_i(S;V).
\end{equation*}

A fair attribution of payouts is considered to satisfy the properties of Efficiency, Symmetry, Dummy and Additivity, and the Shapley value is the only attribution method that has those four properites~\cite{molnar2020interpretable}. More precisely,
\paragraph{Efficiency} All the individual rewards sum to the overall net payouts, i.e., $\sum_i \phi_i = V(\mathcal{N})-V(\emptyset)$, where $V(\emptyset)$ indicates the baseline reward without any player.
\paragraph{Symmetry} If two palyers $i,j$ contribute equally to all possible coalitions, i.e., $V(S\cup \{i\}) = V(S\cup \{j\}), \forall S\subseteq \mathcal{N}/\{i,j\}$, then they have the same reward, $\phi_i = \phi_j$. 
\paragraph{Dummy} A dummy player $i$, which does not contribute to any coalition, i.e., $V(S\cup \{i\}) = V(S)+V(\{i\}), \forall S\subseteq \mathcal{N}/\{i\}$, has the reward as if playing alone, $\phi_i = V(\{i\})$.
\paragraph{Additivity} Consider a combined game $U(S)=V(S)+W(S), \forall S\subseteq \mathcal{N}$, its reward is the sum of the rewards for individual games, i.e. $\phi_i(U) = \phi_i(V) + \phi_i(W)$.

Those four properties in turn could be viewed as axioms defining a fair attribution method, and it has been proved that they uniquely lead to the definition of the Shapley value~\cite{weber1988probabilistic}.

\section{Method}

\subsection{Models for Comparison} 

\begin{figure}[!t]
    \centering
    \includegraphics[width=0.95\columnwidth]{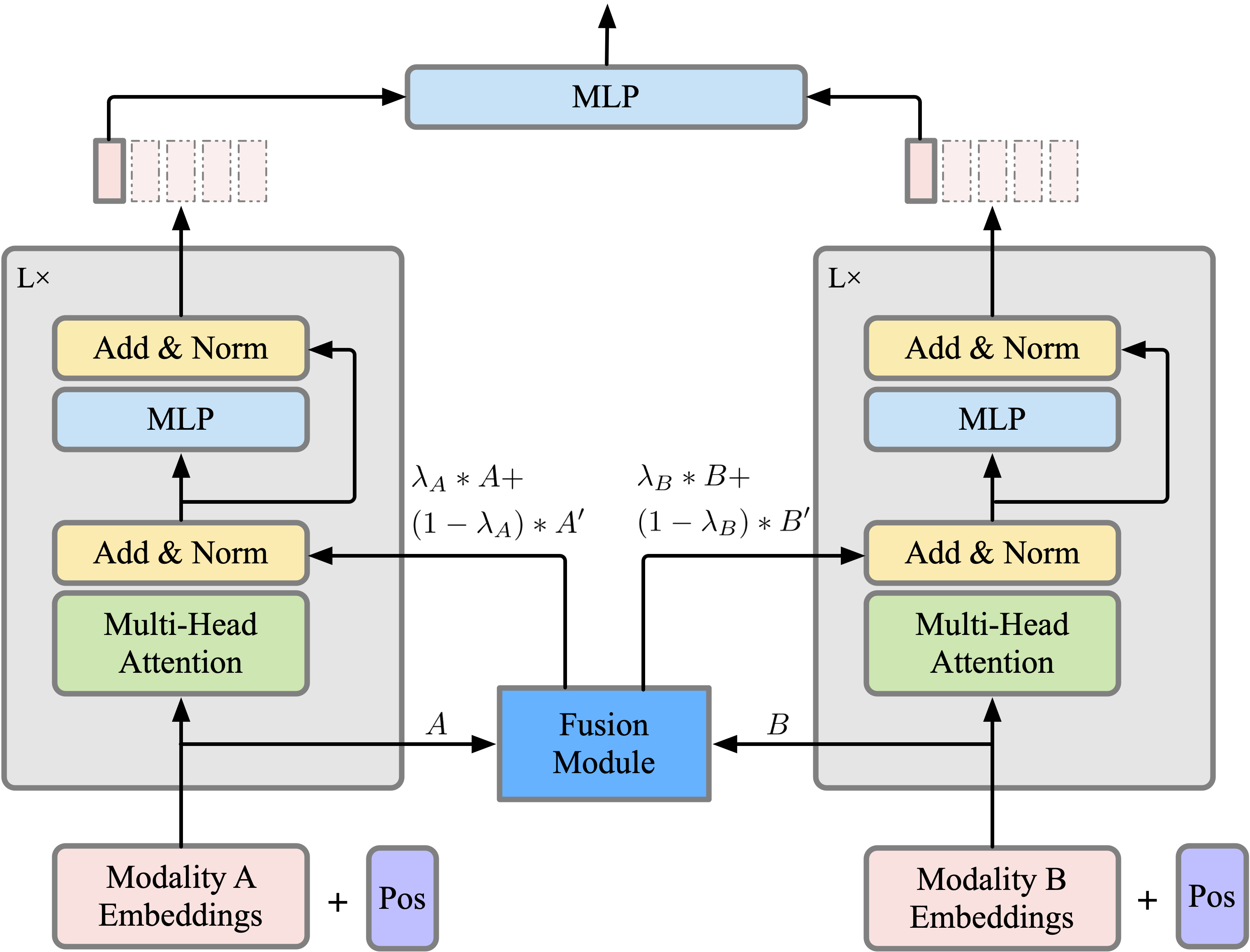}
    \caption{\label{fig:decent}
        Schematic diagram for the dense fusion strategy, implemented with the transformer encoder blocks.
    }
\end{figure}

In deep learning,
diverse fusion methods have been developed to address the multi-modal learning tasks~\cite{baltruvsaitis2018multimodal,wang2021survey}. 
Among them, the Early Fusion and Late Fusion strategies are widely recognized. Besides these two, the rest methods cannot be easily categorized. 
Depending on specific implementation, the fusion may happen at one particular stage or at multiple stages.
In addition, different works usually adopt distinct network architectures, making it infeasible to compare them on the same ground.

The present work is devoted to evaluate the contribution and cooperation of individual modalities under different fusion methods on different multi-modal datasets for different tasks. 
To achieve this goal, we need (i) a common network architecture that ensures decent performance on different datasets for different tasks. 
(ii) a unified framework to account for different fusion types.

To address point (i), we adopt the transformer~\cite{vaswani2017attention} architecture, which is capable of handling different modalities. We are aware that using a common
architecture for different tasks would potentially degrade the performance. 
For point (ii), we introduce the Dense Fusion strategy that builds upon
the transformer architecture (see~\Cref{fig:decent}). The fusion happens at every stage. We also consider one of its variations, which includes weights ($\lambda$'s) to control the fusion intensity. Those weights are learned during the training. Such a variation is referred as the Dynamic Dense Fusion.

To summarize, in our experiments, we consider the Early Fusion, Late Fusion, as well as two variations of the Dense Fusion strategy. All the network models are built upon the transformer architecture. Details can be found in the experiment setup and in the supplementary material.

\subsection{The SHAPE Scores}

The SHAPE scores are built upon the Shapley value with identifying the set of modalities as players and model accuracy as the utility function.
We distinguish two types of SHAPE scores, one captures the marginal contribution of individual modalities, and the other captures the cross-modal cooperation.
The SHAPE scores are easy to compute under the general multi-modal learning scenario, no matter how many modalities are involved.

In the following, we use $V_f(\mathcal{D})$ to denote the accuracy of model $f$ on the dataset $\mathcal{D}$. When discussing the contribution of different modalities, e.g., $M_1$ and $M_2$, we shall slightly change the symbol to $V_f(M_1, M_2)$ for emphasizing the involved modalities. The set of modalities in $\mathcal{D}$ is denoted as $\mathcal{M}:=\{M_1,\dots, M_m\}$.

\subsubsection{SHAPE Score for Modality's Marginal Contribution}

For the purpose of evaluating the marginal contribution of a modality $M_i$, we use the scaled Shapley value 
\begin{equation}\label{marginal}
    \mathcal{S}_{M_i;f;\mathcal{D}} := \frac{1}{Z_f}\phi_{M_i}(\mathcal{M}; V_f),
\end{equation}
where the normalization factor $Z_f$ is introduced to align the scores for different models and thus enable cross-model comparison. Similar to the modal-normalization in~\cite{gat2021perceptual}, we choose $Z_f$ to be the model accuracy on the full dataset $\mathcal{D}$, i.e., $Z_f = V_f(\mathcal{D})$.
For simplicity, we will use the abbreviated symbol $\mathcal{S}_{M_i}$,
when there is no risk of confusion.

As an example, for the two-modal case, we have
\begin{equation*}
    \begin{split}
        \mathcal{S}_{M_1} 
        &= \frac{1}{2Z_f} [V(M_1, M_2) - V(0_1, M_2) \\
        &\qquad+ V(M_1, 0_2) - V(0_1, 0_2)],
    \end{split}
\end{equation*}
where $Z_f = V(M_1, M_2)$. Extensions to three or more modalities cases are straightforward, and we leave them to the supplementary material.

\paragraph{Baseline Values}
The above $0_i$ refers to the absence of modality $M_i$ and should be implemented using some pre-defined baseline values.
In the case where only one modality is absent, it is natural to substitute zeros for the missing inputs.
However, when both modalities are absent, i.e., $V(0_1, 0_2)$, an all-zero input would produce a single deterministic prediction 
that is sensitive to specific training settings, such as the model initialization.
We argue that it is better to predict the majority class in this case. This majority-classifier is stable and could account for the possible unbalances in the dataset.

\subsubsection{SHAPE Score for Cross-modal Cooperation}

The Shapley value also enables us to directly measure the cross-modal cooperation, which evaluates the degree to which the model exploits the complementary information of a modality provided to the other modalities.
% This serves as a quantitative measure of the degree to which the model exploits the complementary information of a modality provides to the other modalities.

Intuitively, 
to compute the cooperation among a set of modalities $A\subseteq \mathcal{M}$, we need to subtract the marginal contributions of all $M_i \in A$ from the marginal contribution of $A$ itself.
Treating $A$ as a single effective player, its marginal contribution can be evaluated by $\phi_A(\mathcal{M}; V_f)$. On the other hand, the marginal contribution of $M_i \in A$ should be $\phi_{M_i}(\mathcal{M}/A\cup M_i; V_f)$. Consequently, the cooperation score is defined as
\begin{equation}
    \mathcal{C}_{\mathcal{A};f;\mathcal{D}} = \phi_A(\mathcal{M}; V_f) - \sum_{M_i\in \mathcal{A}} \phi_{M_i}(\mathcal{M}/A\cup M_i; V_f).
\end{equation}
Note that the set of "players" in the above $\phi_{M_i}$ is different from the one in~\Cref{marginal}. For simplicity, we usually omit the subscripts ${f;\mathcal{D}}$ and write the score as $\mathcal{C}_{\mathcal{A}}$.

For a concrete example, let's consider the two-modal case. After a few steps of simple algebra, we arrive at
\begin{equation*}
    \begin{split}
        \mathcal{C}_{\{M_1,M_2\}} &= V(M_1, M_2) - V(M_1, 0_2) - V(0_1, M_2)\\ 
        &+ V(0_1, 0_2),
    \end{split}
\end{equation*}
where $0_1$ and $0_2$ denote the baseline values for modalities $M_1, M_2$, respectively.

\subsection{Fine-grained Perceptual Scores\label{fgps}}

In this subsection, we proceed to have a closer look at the perceptual score~\cite{gat2021perceptual}.
First, we will show that this score implicitly measures the correlation among different modalities within the dataset $\mathcal{D}$.
Then we propose a fine-grained version of the perceptual score.
In the following, we only keep the key steps and leave the
detailed deduction to the supplementary material.

Let $(x,y)$ denote a data sample in $\mathcal{D}$, where $x \in\mathcal{X}$ is the input feature and $y\in \{0,\dots,k-1\}$ is the corresponding label.
In order to explicitly refer the modalities, we will decompose $x$ as $(x^{\mathcal{M}_1},x^{\mathcal{M}_2})$, where $\mathcal{M}_1 \subseteq \mathcal{M}$ and $\mathcal{M}_2 = \mathcal{M}/\mathcal{M}_1$ are the subsets of modalities.
Recall that the model accuracy is defined as
$V_f(\mathcal{D})=\mathbb{E}_\mathcal{D}[\mathbb{I}_{y^*=y}]$,
where $\mathbb{I}$ is the identity function and $y^*$ refers to the model prediction at input $x$. Simple algebra will lead us to
\begin{equation*}
    V_f(\mathcal{D}) = \int_{\mathcal{X}} dx^{\mathcal{M}_1}dx^{\mathcal{M}_2} \text{Pr}(y*, x^{\mathcal{M}_1}, x^{\mathcal{M}_2}),
\end{equation*}
where $\text{Pr}(y*, x^{\mathcal{M}_1}, x^{\mathcal{M}_2})$ is the probability distribution on $\mathcal{D}$.

The permutation-based step in the perceptual score would alter the data distribution,
resulting in a new effective dataset with distribution $\text{Pr}(y, x^{\mathcal{M}_1})\text{Pr}(x^{\mathcal{M}_2})$.
We refer to this effective dataset as $\mathcal{D}^p$. 
Now, the perceptual score for $\mathcal{M}_2$ is
\begin{equation*}
    \begin{split}
        P_{\mathcal{M}_2;f;\mathcal{D}} &= \frac{1}{Z_f}(V_f(\mathcal{D}) - V_f(\mathcal{D}^p))\\
        &= \frac{1}{Z_f}\int_{\mathcal{X}} dx^{\mathcal{M}_1}dx^{\mathcal{M}_2} P(x^{\mathcal{M}_2})\Delta_{\mathcal{M}_1, \mathcal{M}_2;f},
    \end{split}
\end{equation*}
where $\Delta_{\mathcal{M}_1, \mathcal{M}_2;f} = P(y^*, x^{\mathcal{M}_1}| x^{\mathcal{M}_2}) - P(y^*, x^{\mathcal{M}_1})$. 
Clearly, apart from the dependence on the model $f$ through $y^*$, the term $\Delta_{\mathcal{M}_1, \mathcal{M}_2;f}$ effectively measure the correlation between $\mathcal{M}_1$ and $\mathcal{M}_2$ in the dataset $\mathcal{D}$.

Since $P(x^{\mathcal{M}_2})$ is non-negative, if the part $x^{\mathcal{M}_2}$ comes from the same class as $x^{\mathcal{M}_1}$, then we would expect $\Delta_{M_1, M_2;f}\ge 0$. We refer to this case as the in-class permutation. Likewise, in the out-class permutation, $x^{\mathcal{M}_2}$ comes from a different class from $x^{\mathcal{M}_1}$ and we expect $\Delta_{M_1, M_2;f}\le 0$. 

The in-class and out-class cases provide conflicting driving forces that would cancel each other and 
cannot be reflected by the original perceptual score. 
Based on this observation, we propose to use the in-class and out-class perceptual scores in place of the original one. They are defined as following:
\begin{equation*}
    P^{in/out}_{\mathcal{M}_i;f;\mathcal{D}} := \frac{1}{Z_f}\left[ V_f(\mathcal{M}) - V_f([\mathcal{M}/\mathcal{M}_i]^\pm) \right],
\end{equation*}
where the $[\mathcal{M}/\mathcal{M}_i]^\pm$ operator removes the $\mathcal{M}_i$ modality either by the in-class permutation ($+$) or the out-class permutation ($-$). $Z_f$ is the normalization factor as usual. We only consider the model-normalization here.
Finally, for simplicity, we usually omit the subscripts $f;\mathcal{D}$ as long as it raises no confusion.
\section{Experiments}

\begin{table*}[!ht]
    \centering
    % \small
    \setlength{\tabcolsep}{1.mm}
    \begin{tabular}{lcccccccccccc}
        \toprule
        {B-T4SA} & {} & & \multicolumn{3}{c}{SHAPE scores} & & \multicolumn{6}{c}{(fine-grained) perceptual scores}\\
        \cmidrule{4-6}\cmidrule{8-13}
        {} & Acc. & & $\mathcal{S}_V$ & $\mathcal{S}_T$ & $\mathcal{C}_{VT}$ & & $P_V$ & $P_V^{in}$ & $P_V^{out}$ & $P_T$ & $P_T^{in}$ & $P_T^{out}$ \\
        \midrule
        Unimodal V     &  41.95  & &    --     &    --       &    --      & &     --             &     --             &     --             &     --             &     --             &     --             \\
        Unimodel T     &  94.89  & &    --     &    --       &    --      & &     --             &     --             &     --             &     --             &     --             &     --             \\
        Early Fusion   &  95.01  & &   $0.52$  &   $64.40$   &   $1.04$   & &   $0.02$ {\tiny $\pm 0.01$}   &   $ 0.03$ {\tiny $\pm 0.01$}  &   $0.03$ {\tiny $\pm 0.02$}   &   $65.00$ {\tiny $\pm 0.26$}  &  $-0.04$ {\tiny $\pm 0.14$}   &  $97.34$ {\tiny $\pm 0.09$}   \\
        Late Fusion    &  95.22  & &   $0.00$  &   $65.00$   &   $0.00$   & &   $2.09$ {\tiny $\pm 0.01$}   &   $-0.01$ {\tiny $\pm 0.01$}  &   $0.61$ {\tiny $\pm 0.01$}   &   $64.95$ {\tiny $\pm 0.27$}  &  $-0.02$ {\tiny $\pm 0.11$}   &  $97.43$ {\tiny $\pm 0.12$}   \\
        Dense Fusion &  94.97  & &   $0.24$  &   $64.66$   &   $0.48$   & &   $0.01$ {\tiny $\pm 0.01$}   &   $ 0.01$ {\tiny $\pm 0.01$}  &   $0.02$ {\tiny $\pm 0.01$}   &   $64.84$ {\tiny $\pm 0.24$}  &  $ 0.14$ {\tiny $\pm 0.13$}   &  $97.23$ {\tiny $\pm 0.09$}   \\
        Dynamic Fusion  &  95.36  & &   $0.01$  &   $65.04$   &   $0.01$   & &   $0.01$ {\tiny $\pm 0.01$}   &   $ 0.01$ {\tiny $\pm 0.01$}  &   $0.00$ {\tiny $\pm 0.01$}   &   $65.02$ {\tiny $\pm 0.30$}  &  $ 0.05$ {\tiny $\pm 0.14$}   &  $97.57$ {\tiny $\pm 0.17$}   \\
        \bottomrule
    \end{tabular}
    \caption{\label{tab:t4sa}
        Accuracy (Acc.), 
        the SHAPE scores ($\mathcal{S}_V$, $\mathcal{S}_T$, $\mathcal{C}_{VT}$)
        and the (fine-grained) perceptual scores ($P_{V/T}$, $P_{V/T}^{in}$, $P_{V/T}^{out}$) 
        on the B-T4SA dataset for multi-modal models using different fusion strategies.
        Dynamic Fusion stands for the Dynamic Dense Fusion.
        The results of unimodal models are included as baselines.
        $V$ and $T$ indicate the visual and text modalities, respectively.
    }
\end{table*}
\begin{table*}[!ht]
    \centering
    % \small
    \setlength{\tabcolsep}{0.7mm}
    \begin{tabular}{lcccccccccccc}
        \toprule
        {CMU-MOSEI} & {} & & \multicolumn{3}{c}{SHAPE scores} & & \multicolumn{6}{c}{(fine-grained) perceptual scores}\\
        \cmidrule{4-6}\cmidrule{8-13}
        {} & Acc. & & $\mathcal{S}_A$ & $\mathcal{S}_T$ & $\mathcal{C}_{AT}$ & & $P_A$ & $P_A^{in}$ & $P_A^{out}$ & $P_T$ & $P_T^{in}$ & $P_T^{out}$ \\
        \midrule
        Unimodal A     &  71.04  & &    --     &    --     &    --    & &     --             &     --             &     --             &     --             &     --             &     --             \\
        Unimodel T     &  80.20  & &    --     &    --     &    --    & &     --             &     --             &     --             &     --             &     --             &     --             \\
        Early Fusion   &  80.22  & &   -0.01   &   11.46   &  -0.02   & &  $ 1.21$ {\tiny $\pm 0.17$}   &   $-0.62$ {\tiny $\pm 1.50$}  &   $-0.30$ {\tiny $\pm 0.16$}  &   $21.76$ {\tiny $\pm 0.64$}  &   $-0.47$ {\tiny $\pm 0.24$}  &  $59.85$ {\tiny $\pm 1.66$}  \\
        Late Fusion    &  80.67  & &    0.04   &   11.89   &   0.09   & &  $ 0.72$ {\tiny $\pm 0.09$}   &   $ 2.12$ {\tiny $\pm 0.22$}  &   $ 2.38$ {\tiny $\pm 0.31$}  &   $22.04$ {\tiny $\pm 0.76$}  &   $ 2.91$ {\tiny $\pm 1.23$}  &  $50.68$ {\tiny $\pm 1.45$}  \\
        Dense Fusion   &  79.04  & &    0.19   &    9.93   &   0.38   & &  $-0.08$ {\tiny $\pm 0.20$}   &   $ 0.65$ {\tiny $\pm 0.47$}  &   $ 0.53$ {\tiny $\pm 0.33$}  &   $19.04$ {\tiny $\pm 0.90$}  &   $ 0.90$ {\tiny $\pm 1.19$}  &  $46.95$ {\tiny $\pm 1.28$}  \\
        Dynamic Fusion &  80.05  & &   -0.21   &   11.47   &  -0.42   & &  $ 2.41$ {\tiny $\pm 0.14$}   &   $ 0.47$ {\tiny $\pm 0.17$}  &   $ 0.65$ {\tiny $\pm 0.22$}  &   $21.99$ {\tiny $\pm 0.87$}  &   $ 0.31$ {\tiny $\pm 0.95$}  &  $54.60$ {\tiny $\pm 1.75$}  \\
        \bottomrule
    \end{tabular}
    \caption{\label{tab:mosei}
        The same as~\Cref{tab:t4sa}, but for the MOSEI dataset.
        The invoved modalities are audio (A) and text (T).
    }
\end{table*}
\begin{table*}[!ht]
    \centering
    % \small
    \setlength{\tabcolsep}{0.7mm}
    \begin{tabular}{lcccccccccccc}
        \toprule
        {SNLI} & {} & & \multicolumn{3}{c}{SHAPE scores} & & \multicolumn{6}{c}{(ffine-grained) perceptual scores}\\
        \cmidrule{4-6}\cmidrule{8-13}
        {} & Acc. & & $\mathcal{S}_{T_p}$ & $\mathcal{S}_{T_h}$ & $\mathcal{C}_{T_pT_h}$ & & $P_{T_p}$ & $P_{T_p}^{in}$ & $P_{T_p}^{out}$ & $P_{T_h}$ & $P_{T_h}^{in}$ & $P_{T_h}^{out}$ \\
        \midrule
        Unimodal $T_p$  &  33.84  & &    --      &    --     &    --     & &     --            &     --             &     --             &     --             &     --             &     --             \\
        Unimodel $T_h$  &  66.32  & &    --      &    --     &    --     & &     --            &     --             &     --             &     --             &     --             &     --             \\
        Early Fusion    &  82.47  & &  $21.34$   &  $38.25$  &  $42.52$  & &  $42.89$ {\tiny $\pm 0.52$}  &  $42.89$ {\tiny $\pm 0.30$}   &   $42.50$ {\tiny $\pm 0.15$}  &  $59.84$ {\tiny $\pm 0.25$}   &   $42.42$ {\tiny $\pm 0.76$}  &   $68.53$ {\tiny $\pm 0.32$}  \\
        Late Fusion     &  69.77  & &  $ 6.41$   &  $45.82$  &  $10.55$  & &  $ 6.97$ {\tiny $\pm 0.40$}  &  $ 6.94$ {\tiny $\pm 0.30$}   &   $ 7.30$ {\tiny $\pm 0.47$}  &  $52.33$ {\tiny $\pm 1.18$}   &   $ 7.07$ {\tiny $\pm 0.79$}  &   $60.69$ {\tiny $\pm 0.46$}  \\
        Dense Fusion &  84.91  & &  $25.49$   &  $35.26$  &  $50.95$  & &  $47.65$ {\tiny $\pm 0.28$}  &  $47.73$ {\tiny $\pm 0.37$}   &   $47.58$ {\tiny $\pm 0.18$}  &  $61.04$ {\tiny $\pm 0.25$}   &   $47.50$ {\tiny $\pm 0.62$}  &   $67.87$ {\tiny $\pm 0.45$}  \\
        Dynamic Fusion   &  85.02  & &  $25.61$   &  $35.19$  &  $52.06$  & &  $48.40$ {\tiny $\pm 0.25$}  &  $48.24$ {\tiny $\pm 0.31$}   &   $48.66$ {\tiny $\pm 0.38$}  &  $61.06$ {\tiny $\pm 0.47$}   &   $48.09$ {\tiny $\pm 0.68$}  &   $67.54$ {\tiny $\pm 0.49$}  \\
        \bottomrule
    \end{tabular}
    \caption{\label{tab:snli}
        The same as~\Cref{tab:t4sa}, but for the SNLI dataset.
        The invoved modalities are the premise text ($T_p$) and the hypothesis text ($T_h$).
    }
\end{table*}

\subsection{Setup} 
For all models, we use 4-layer transformer encoder as the backbone. 
The number of attention heads is 8, and the hidden dimension is 1024. 
All models are trained using the Adam optimizer with learning rate warm-up
in the first epoch.
We encode text data with the GloVe~\cite{pennington2014glove}, 
and audio data with the MFCC~\cite{tiwari2010mfcc}. 
As for the image data, 
we employ the patch embedding technique, following
the ViT Model~\cite{dosovitskiy2020image}. 
It is worth noting that our models achieve comparable performance with the supervision-based SOTA models.

\subsection{Datasets}

Real-life applications raise diverse demands for multi-modal data. Depending on specific tasks, some must be handled by combining all the involved modalities, while in the others, additional modalities merely provide complementary information. 
Furthermore, different modalities in a multi-modal dataset might be only weakly correlated to each other, or they could be different representations that provide precisely the same information. 
To account for those complicated situations, we deliberately choose three datasets, as will be introduced below.

\paragraph{B-T4SA} The dataset~\cite{vadicamo2017cross} is collected from Twitter, 
including the posted text (textual modality T) and associated images (visual modality V). Each data pair is annotated with a sentiment label (positive, negative and neutral).
The task is to predict this label using both the text and image inputs.
We note that the text data normally align with their sentiment labels while
a large number of images seem to be irrelevant to the labels. 
B-T4SA represents the situation that different modalities are complementary and weakly correlated to each other.

\paragraph{CMU-MOSEI} This dataset~\cite{zadeh2018multimodal} is the largest dataset for sentence-level sentiment analysis and emotion recognition. It contains more than 65 hours of annotated video data, consisting of audio, video and text modalities. In our experiments, we only use the audio (A) and text (T) modalities. The text modality includes the scripts that are generated from the audio data. In this way, the CMU-MOSEI dataset stands for the situation that different modalities are complementary and highly correlated.

\paragraph{SNLI} The SNLI dataset~\cite{bowman2015large} is a collection of 570k human-written English sentence pairs. Each pair contains a premise and a hypothesis, whose relation is labeled as either entailment, contradiction, or neutral. 
We shall view the premise and hypothesis as two distinct modalities, i.e., $T_{p}$ and $T_{h}$.
The underlying task is to infer the relationship between the premise and the hypothesis. Clearly, this task intrinsically relies on both modalities and, thus, represents the situation that different modalities are indispensable.

\subsection{Contribution and Cooperation}

In this subsection, we investigate
the SHAPE scores across different models on the three multi-modal datasets, demonstrating their ability in capturing the marginal contribution of individual modalities and the cross-modal cooperation.

% T4SA
\Cref{tab:t4sa} summarizes the results for B-T4SA dataset.
We find the unimodal model for T performs as good as multi-modal models, while the unimodal model for V only slightly suppasses the random guess. 
The SHAPE scores align with these observations.
The contribution scores $\mathcal{S}_V$ for V modality are near zero for all the
fusion strategies, indicating V almost does not contribute to the task.
Further, the cooperation scores $\mathcal{C}_{VT}$ are very small as well.
Hence, as expected, nearly no cooperation happens between the two modalities.
At last, we note that $\mathcal{C}_{VT}$ for early fusion and dense fusion are evidently higher than the other two models.
This may be attributed to the fact that both the early and the dense fusion models encourage the fusion between the two modalities. 

% MOSEI
The A modality in the CMU-MOSEI dataset encodes richer information than its text counterparts by including features such as the tone and the talking speed. On the other hand, the T modality is better at representing semantic information. 
One would expect a multi-modal model to actively integrate the two modalities and make better predictions. Unfortunately, it is not the case as indicated by the SHAPE scores (see~\Cref{tab:mosei}). Values of $\mathcal{S}_A$ and $\mathcal{C}_{AT}$ are very small, indicating the audio data seldom contribute to the model performance and there is little cooperation between the A and T modalities.
In addition, the marginal contributions of T modality, $\mathcal{S}_T$, are also relatively small. At first sight, it may seem contradictory that both modalities have limited contribution to the performance, while the model achieves a decent accuracy.
Closer investigation reveals that this is a consequence of the unbalanced dataset.
Simply by predicting positive, one could get an accuracy of $71.04\%$, which is the
accuracy of the unimodal model for A.
This phenomenon coincides with the common belief that the deep neural networks are lazy, tending to take shortcuts. It could be the case that the multi-modal models simply exploit the unbalance of the dataset and pay little effort in mining the cooperation.

% SNLI
In~\Cref{tab:snli}, we show the results for the natural language inference task on
the SNLI dataset,
for which both modalities are indispensable.
We find that both unimodal models perform poorly. 
Interestingly, the unimodal model for $T_h$
significantly outperforms the random guess. This implies that there could be implicit biases within the dataset despite that it is balanced. 
As expected, the behavior of SHAPE scores is different from previous cases.
$\mathcal{S}_{T_p}$ and $\mathcal{S}_{T_h}$ becomes comparable, with $\mathcal{S}_{T_h}$ being generally higher.
Hence, both modalities contribute, and the hypothesis is more important for model prediction.
More strikingly, the cooperation scores $\mathcal{C}_{{T_p}{T_h}}$ are also high.
This suggests that the models actively exploit the cooperation between $T_p$ and $T_h$ for predictions.
At last, in the late fusion case, 
$\mathcal{S}_{T_p}$ and $\mathcal{C}_{{T_p}{T_h}}$ are much lower than the other cases. 
The lack of cooperation could be attributed to the nature of the late fusion strategy, i.e., the fusion only happen at the last step.
This helps explain the low accuracy of the late fusion model and suggest that, in this case, fusion at earlier stages is a better choice.

\subsection{Compared with the Perceptual Score}

In this subsection, we compare our SHAPE scores with the perceptual score as well as with its fine-grained versions. 

Firstly, we observe that the perceptual score $P$ coincides with our marginal contribution SHAPE score $\mathcal{S}$
in the sense that their values are highly correlated.
In the B-T4SA dataset, both $\mathcal{S}_V$ and $P_V$ are close to zero, and $\mathcal{S}_T$ ($P_T$) is significantly higher than $\mathcal{S}_V$ ($P_V$). Similar pattern is found in the CMU-MOSEI dataset.
For the SNLI dataset, $\mathcal{S}$ and $P$ are both considerably higher than zero, 
except for the Late Fusion case. 
Besides, both scores for $T_h$ modality are higher than the ones for  $T_p$ modality.

Despite the similarities, 
% as mentioned above, 
we emphasize the superiority of our SHAPE scores for 
(i) guaranteed by the game theory, the SHAPE score $\mathcal{S}$ can correctly capture the marginal contribution no matter how many modalities are involved. 
(ii) the SHAPE score $\mathcal{C}$ can directly quantify the cooperation among modalities in addition to their individual importance.

Next, we turn to the fine-grained perceptual scores. As argued in~\cref{fgps}, the raw perceptual score implicitly measures the correlation among modalities in the dataset. However, the positive and negative correlations are mixed and cancel each other, causing a considerable loss of information.
To be concrete, let's consider the Late Fusion case for the SNLI dataset. With only the raw perceptual scores, we at best might conclude that the $T_h$ modality is more important than the $T_p$ one in the contribution to model performance.
On the other hand, we find $P^{in}_{T_h}$ is much smaller than $P^{out}_{T_h}$, indicating that the model prediction almost relies solely on the $T_h$ modality. 
The small values of $P^{in}_{T_p}$ and $P^{out}_{T_p}$ further consolidate this conclusion. Putting it differently, the Late Fusion model tends to use only the $T_h$ modality while ignoring the $T_p$ modality. Consequently, the cooperation between the two modalities should be small in the Late Fusion case.
This is just what the cooperation SHAPE score $C_{T_pT_h}$ reveals to us.
Under more complicated scenario with three or more modalities, it would be harder to capture the cross-modal cooperation using perceptual scores. 
On the contrary, it is always straightforward to quantify the cross-modal cooperation with the SHAPE score $\mathcal{C}$.

\section{Discussion}

The above experiment results suggest that the way a model exploits multi-modal information depends on various factors, including the intrinsic correlation among individual modalities in the dataset, the type of the task, and the fusion strategy in use. 
Based on this observation, we want to draw attention to 
% the importance of choosing 
dataset and task in multi-modal learning research. There is little hope to train a model that can exploit cross-modal cooperation on a dataset where different modalities are nearly independent. 
Likewise, if the task can be easily handled with only a single modality, one must try hard to design an algorithm to exploit the possible cross-modal cooperation.
Furthermore, the bias and unbalance in dataset also affect multi-modal learning. 
In many cases, the improvement in the accuracy of a multi-modal model could be illusive. 
Our SHAPE score $\mathcal{C}$ can directly quantify the cooperation among individual modalities and, hence, could help diagnose the fusion strategy.

At last, it is worth noting that though the present SHAPE scores are defined using the accuracy metric as the utility function, 
% we would like to emphasize that 
there is no limitation of what function should be used. 
In fact, the SHAPE scores can readily adapt to more complex metrics such as the F1 score.

\section{Conclusion}
In this paper, we present the SHAPE scores that quantify the marginal contribution of individual modalities and their collaboration. 
Properties of the Shapely value provides theoretical guarantee for the validity of our SHAPE scores and the experiments empirically verify their effectiveness.
We would like to suggest using the SHAPE scores in addition to the accuracy metric when analyzing a multi-modal model, 
as they are easy to compute and can extract key information about how the model use multi-modal data.

We are aware that the present SHAPE scores are not yet perfect. The choice of baseline values deserve more deliberate consideration. 
In the future work, we will keep improving the SHAPE scores and systematically evaluate the SOTA multi-modal models.

\section*{Acknowledgments}

This work was supported by grants from the National Science and Technology Innovation 2030 Project of China to Yi Zhou (2021ZD0202600).

% reference
\bibliography{SHAPE}
\bibliographystyle{icml2021}

%%%%%%%%%%%%%%%%%%%%%%%%%%%%%%%%%%%%%%%%%%%%%%%%%%%%%%%%%%%%%%%%%%%%%%%%%%%%%%%
% Appendix
%%%%%%%%%%%%%%%%%%%%%%%%%%%%%%%%%%%%%%%%%%%%%%%%%%%%%%%%%%%%%%%%%%%%%%%%%%%%%%%
\appendix
\onecolumn
\section{Appendix}

\subsection{Model Details}
\paragraph{Setting} Here we define the A,B as the input embedding of two modalities. Moreover, all models use a four-layer transformer encoder module as the backbone, including both the unimodal model and multi-modal modal. The encoder module can be described as follows:

\begin{itemize}
    \item Component 1. 
    A  Multi-Head  Self-attention module with a residual operation:  $LayerNorm (X+ Attention (X))$
    \item Component 2.
    A Feedforward module with a residual operation $LayerNorm(X+ FFNN(X))$ , where FFNN function is calculated by $FFNN(X) = W_2(W_1X) $  
    \item Component 3. 
    A MLP module that consists of only a one-layer Linear module, which projects the dimension of the hidden layer to the dimension of class numbers.
    
\end{itemize}
In addition, we add a learnable $CLS$ token to the sequence of embedded features ( The function is similar to the BERT [class] token ), and a position embedding is added to the embedded features. 

\subsubsection{Early Fusion Model}
In the Early fusion model, we simply implement the $Concat$ operation on each modality's embedding feature: $[A, B] = CONCAT (A, B)$. Then we combined embedding$ [A, B]$ as the input to a one pathway transformer encoder-based model. 

\subsubsection{Late Fusion Model}
In the Late fusion model, we first use two pathway structures to learn the representations of individual modalities independently, and then we combine the first vector of the last output hidden layer from each pathway to be the final output. The model's architecture is the same as Figure 1 but does not have the fusion model. 
\subsubsection{Dense Fusion Model}
In the dense fusion model, we integrate modalities using a Multi-head Self-attention module to learn co-representation between two modalities. The operation consists of two steps:
    \begin{itemize}
        \item[1.] 
        We get the co-representation $A' $and $B' $ by  $A', B'= Attention(CONCAT (A, B))$.  
        \item[2.] 
        We make modalities fusion operation by  $LayerNorm (A'+ Attention(A))$ and $LayerNorm (B'+ Attention(B))$ which allows the model to fuse the co-representation with individual modalities representation.  We implement this operation on every layer of the transformer encoder model. 
        
    \end{itemize}
       
\subsubsection{Dynamic Dense Fusion Model}
The architecture of this model is similar to the Dense Fusion Model, but we add a learned parameter $\lambda$ to dynamic control how much the co-representation can be integrated into each modalities' pathway. The process can be described as follows: 
    \begin{itemize}
        \item[1.] 
        We get the co-representation $A' $and $B' $ by  $A', B'= Attention (CONCAT (A, B))$.  
        \item[2.] 
        We get the parameter $\lambda$ by a Gate Function which can be wrote by pytorch-like code as follows:  
        \begin{equation*}
        Gate(A,B) = nn.Sigmoid(nn.Linear(nn.Flatten(nn.MaxPooling(nn.Linear(CONCAT(A,B)))),num)),
        \end{equation*}
        where the $num$ is the number of $\lambda$ that needs to be learned in our multi-modal models, which also means the number of the modalities used in our model.  
        We make modalities fusion operation by$LayerNorm (\lambda_{A} * A + (1-\lambda_A)*A'+ Attention(A))$ and $LayerNorm (\lambda_{B} * B + (1-\lambda_B)*B'+ Attention(B))$.
    \end{itemize}

\section{Deduction Detail}

\subsection{Example of the SHARP Score $\mathcal{S}$ for Three-modality Case}

Here we provide an concrete example of the marginal contribution in the three-modality case, 
i.e., $\mathcal{M} = \{{{M}_1,{M}_2,{M}_3} \}$
\begin{align*}
 \mathcal{S}_{{M}_1;f;\mathcal{D}}&= \frac{1}{{Z}_f}* \{ \frac{2}{6}[V_f({M}_1,{M}_2,{M}_3 - V_f(0_1,{M}_2,{M}_3)] \\
&+ \frac{1}{6}[V_f({M}_1,{M}_2,{0}_3)-V_f({0}_1,{M}_2,{0}_3)]  \\
&+ \frac{1}{6}[V_f({M}_1,{0}_2,{M}_3)-V_f({0}_1,{0}_2,{M}_3)]  \\
&+ \frac{2}{6}[V_f({M}_1,{0}_2,{0}_3)-V_f({0}_1,{0}_2,{0}_3)] 
\end{align*}

\subsection{Perceptual Score}
The model accuracy is defined as
\begin{equation*}
    \begin{split}
        {V}_f(\mathcal{D}) &= \mathbb{E}_\mathcal{D}[I_{y^*=y}]\\
        & = \underset{i\in[K]}{\sum}\text{Pr}(y^*=i,y=i),
    \end{split}
\end{equation*}
where $y^* = \underset{y}{argmax}f_y(x,\theta^*)$ is model prediction and $[K]:=\{0,\dots,K-1\}$ refers to the set of possible labels.

We proceed as follows:
\begin{align*}
 V_f(\mathcal{D})&= \underset{i\in[K]}{\sum}\text{Pr}(y^*=i,y=i) \\
&= \int_{\mathcal{X}}dx^{\mathcal{M}_1}dx^{\mathcal{M}_2} \underset{i}{\sum}\text{Pr}(y^*=i,y=i,x^{\mathcal{M}_1},x^{\mathcal{M}_2}) \\
&=  \int_{\mathcal{X}}dx^{\mathcal{M}_1}dx^{\mathcal{M}_2} \underset{i}{\sum}\text{Pr}(y^*=i| y=i,x^{\mathcal{M}_1},x^{\mathcal{M}_2})\text{Pr}(y=i,x^{\mathcal{M}_1},x^{\mathcal{M}_2})\\
&=  \int_{\mathcal{X}}dx^{\mathcal{M}_1}dx^{\mathcal{M}_2} \underset{i}{\sum}\text{Pr}(y^*=i|x^{\mathcal{M}_1},x^{\mathcal{M}_2})\text{Pr}(y=i,x^{\mathcal{M}_1},x^{\mathcal{M}_2})\\
&=   \int_{\mathcal{X}}dx^{\mathcal{M}_1}dx^{\mathcal{M}_2} \text{Pr}(y^*=l^*,x^{\mathcal{M}_1},x^{\mathcal{M}_2})
\end{align*}
where $l^*$ is the predicted label at $x = (x^{\mathcal{M}_1},x^{\mathcal{M}_2})$, and in 
the last step we use following equation,
$$\text{Pr}(y^*=i|x^{\mathcal{M}_1},x^{\mathcal{M}_2}) = \delta_{i,l^*}.$$

The permutation step in the perceptual score will alter the data distribution and results in a new effective dataset $\mathcal{D}^p$.
Suppose we are calculating the perceptual score for modality set $\mathcal{M}_2$, the discribution of $\mathcal{D}^p$ will be 
$\text{Pr}(y,x^{\mathcal{M}_1})\text{Pr}(x^{\mathcal{M}_2})$. Through similar steps, we arrive at
\begin{equation*}
    V_f(\mathcal{D}^p) = \int_{\mathcal{X}}dx^{\mathcal{M}_1}dx^{\mathcal{M}_2} \text{Pr}(y^*=l^*,x^{\mathcal{M}_1})\text{Pr}(x^{\mathcal{M}_2})
\end{equation*}
Then, the perceptual score can be expressed as
\begin{equation*}
    \begin{split}
        P_{\mathcal{M}_2;f;\mathcal{D}} &= \frac{1}{Z_f} (V_f(\mathcal{D})-V_f(\mathcal{D}^p))\\
        &= \frac{1}{Z_f}\int_{\mathcal{X}}dx^{\mathcal{M}_1}dx^{\mathcal{M}_2}
        \text{Pr}(x^{\mathcal{M}_2})
        [P(y^*,x^{\mathcal{M}_1}|x^{\mathcal{M}_2})-P(y^*,x^{\mathcal{M}_1})].
    \end{split}
\end{equation*}

%%%%%%%%%%%%%%%%%%%%%%%%%%%%%%%%%%%%%%%%%%%%%%%%%%%%%%%%%%%%%%%%%%%%%%%%%%%%%%%

\end{document}